# An Improved Residual LSTM Architecture for Acoustic Modeling

Lu Huang
Department of Electronic Engineering
Tsinghua University
Beijing, China
e-mail: huanglu.th@gmail.com

Jiasong Sun
Department of Electronic Engineering
Tsinghua University
Beijing, China
e-mail: sunjiasong@tsinghua.edu.cn

Ji Xu
Key Lab of Speech Acoustics & Content Understanding
Institute of Acoustics, Chinese Academy of Sciences
Beijing, China
e-mail: xuji@hccl.ioa.ac.cn

Yi Yang
Department of Electronic Engineering
Tsinghua University
Beijing, China
e-mail: yangyy@tsinghua.edu.cn

*Abstract*—Long Short-Term Memory (LSTM) is the primary recurrent neural networks architecture for acoustic modeling in automatic speech recognition systems. Residual learning is an efficient method to help neural networks converge easier and faster. In this paper, we propose several types of residual LSTM methods for our acoustic modeling. Our experiments indicate that, compared with classic LSTM, our architecture shows more than 8% relative reduction in Phone Error Rate (PER) on TIMIT tasks. At the same time, our residual fast LSTM approach shows 4% relative reduction in PER on the same task. Besides, we find that all this architecture could have good results on THCHS-30, Librispeech and Switchboard corpora.

*long short-term memory; residual learning; acoustic modeling; automatic speech recognition*

## I. INTRODUCTION

Artificial Neural Networks (ANNs) had been widely researched in many Automatic Speech Recognition (ASR) systems for a long time. Since the year of 2011, Microsoft proposed its first deep neural networks which extremely improved the performance of speech recognition system [1]. After that, Deep Neural Networks (DNNs), Convolution Neural Networks (CNNs) and Recurrent Neural Networks (RNNs) have been the most important research and development ways and tools. But there are still some problems on training deeper network for the existence of vanishing gradients and exploding gradients [2]. The standard LSTM-RNNs [3] has been designed to address these problems [4].

At the same time, the residual networks have been applied in image recognition [5] and speech recognition [6–10]. Especially, residual LSTM is proposed to improve the performance of speech recognition systems [6, 10]. Meanwhile, fast LSTM [11, 12] is proposed to reduce the training time without sacrificing performance, sometimes even with improved performance.

In this paper, inspired by Residual LSTM in [6], we propose an improved residual LSTM to replace vector's addition by splicing vector with various shortcut connection locations. They have good performance for acoustic modeling on TIMIT [13], THCHS-30 [14], Librispeech [15] and Switchboard [16] tasks.

We start by describing some fundamental LSTM architectures in Section II, including standard LSTM, projected LSTM [4], fast LSTM and residual LSTM [6]. Then the improved residual LSTM is proposed in Section III. After that, we provide some experiments and results on TIMIT, THCHS-30, Librispeech and Switchboard corpora in Section IV, which is followed by conclusions in Section V.

## II. FUNDAMENTAL LSTM-RELATED ARCHITECTURES

Since LSTM was proposed in [3], it has achieved significant performance in sequence labelling and prediction [17], especially as acoustic modeling and language modeling in ASR systems.

In this section, some LSTM-related works are provided, included projected LSTM, fast LSTM, and residual LSTM [6].

### A. LSTM

In the year of 2014, the projected LSTM is proposed by Google for speech recognition [4], which is called LSTM projected (LSTMP) architecture in some other literatures. When compared with the standard LSTM, projected LSTM add a projection layer to the output, and also the recurrent vector is truncated from the output vector in Kaldi [18], as illustrated in Figure 1 and shown by the following equations.

$$i_t = \sigma(W_{ix}x_t + W_{ir}r_{t-1} + W_{ic}c_{t-1} + b_i) \quad (1)$$

$$f_t = \sigma(W_{fx}x_t + W_{fr}r_{t-1} + W_{fc}c_{t-1} + b_f) \quad (2)$$

$$g_t = \tanh(W_{gx}x_t + W_{gr}r_{t-1} + b_g) \quad (3)$$

$$c_t = i_t \odot g_t + f_t \odot c_{t-1} \quad (4)$$

$$o_t = \sigma(W_{ox}x_t + W_{or}r_{t-1} + W_{oc}c_t + b_o) \quad (5)$$

$$m_t = o_t \odot \tanh(c_t) \quad (6)$$

$$y_t = W_{rp}m_t \quad (7)$$

$$r_t = y_t(1:n_r) \quad (8)$$

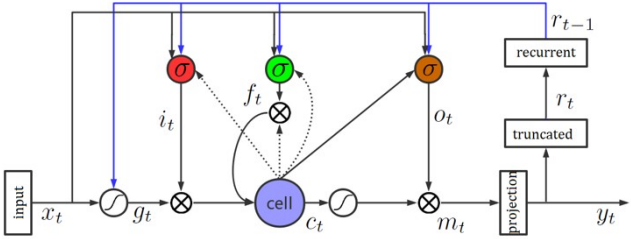

Figure 1. The projected LSTM in Kaldi.

Where the $W$ represents weight matrix and the $b$ is bias vector. For example, $W_{ic}$ is the matrix of weights from cell activation vectors to input gate, $b_i$ is the bias vector of input gate. $i_t$, $f_t$ and $o_t$ are the input gate, forget gate and output gate respectively. $c_t$ and $m_t$ are cell activation vector and cell output activation vector. All of $i_t$, $f_t$, $o_t$, $c_t$ and $m_t$ have the same size. Besides, $g_t$ is the processed input, $r_t$ is for recurrence, $y_t$ is the output, $W_{rp}$ is the projection matrix. $\odot$ stands for element-wise multiplication, $y_t(1:n_r)$ means that $r_t$ is the first $n_r$ elements of $y_t$.

### B. Fast LSTM

LSTM-RNN has more complexity than general RNN, which leads to a slower training speed. To accelerate its training, several algorithms have been done in previous work [11, 12, 19], in which the fast LSTM [11, 12] can cut down half training time with incidentally improvement of performance. Compared to LSTM, fast LSTM doesn't use $c_t$ or $c_{t-1}$ to compute $i_t$, $f_t$ and $o_t$, as shown in the following equations.

$$i_t = \sigma(W_{ix}x_t + W_{ir}r_{t-1} + b_i) \quad (9)$$

$$f_t = \sigma(W_{fx}x_t + W_{fr}r_{t-1} + b_f) \quad (10)$$

$$o_t = \sigma(W_{ox}x_t + W_{or}r_{t-1} + b_o) \quad (11)$$

Obviously, (9), (10), (11) and (3) have the same inner operation, except that the active function is different for (3).

Since these four equations have the same inner operation and inputs, a larger matrix is adopted in Kaldi to convert them into one operation with a large output, which contains four parts to represent the inputs of active functions: $i_t$, $f_t$, $o_t$ and $g_t$. This kind of large matrix is more suitable for Graphics Processing Unit (GPU) computing than small matrices when they have the same number of parameters.

### C. Residual LSTM

As mentioned above, the DNN's training will become harder with the increase of depth. The reason is attributed to vanishing gradients and exploding gradients. Residual learning has been proposed to solve these problems in image recognition [5], and recently in speech recognition [6, 10].

In traditional residual LSTM architecture, the output of each layer is the sum of network's input and network's output in [10], i.e., there is one shortcut connection between network's input and output, which has the same form as on most image recognition tasks. However, the shortcut connection is ended at one network's interior node, instead of network's output in [6].

## III. THE IMPROVED RESIDUAL LSTM ARCHITECTURES

Inspired by the principle in [6], our LSTM architectures have a short connection in LSTM on three different locations. Two of them are inner nodes and one of them is output, which is shown in Figure 2. Besides, we replace vector's addition by splicing vector in each location, and then project the spliced vector to the original dimension in location 1 and 3 to prevent the exploding dimension of vectors. In location 2, we reuse the projection matrix $W_{rp}$ by increasing its input dimension.

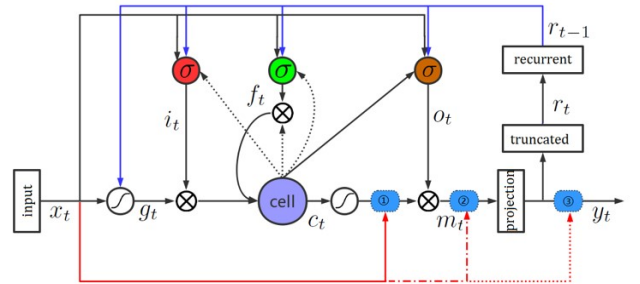

Figure 2. The improved residual LSTM.

When the junction is located at position 1, as shown in Figure 2, we name our improved Residual LSTM as LSTM Res-1, and the same is true for LSTM Res-2 and LSTM Res-3. So, compared to the original LSTM, our first method LSTM Res-1 has different equations from (6) as below.

$$h_t = (\tanh(c_t), x_t) \quad (12)$$

$$m_t = o_t \odot (W_{Res1}h_t) \quad (13)$$

Where $h_t$ is the spliced vector, $W_{Res1}$ is the projection matrix. Our second method LSTM Res-2 has different equations from (7) as below.

$$h_t = (m_t, x_t) \quad (14)$$

$$m_t = W_{Res2} h_t \quad (15)$$

Where $W_{Res2}$ is the projection matrix for replacing $W_{rp}$ in (7). Similarly, the third one LSTM Res-3 changes (7) and (8) into following equations.

$$z_t = W_{rp} m_t \quad (16)$$

$$r_t = z_t(1:n_r) \quad (17)$$

$$h_t = (z_t, x_t) \quad (18)$$

$$y_t = W_{Res3} h_t \quad (19)$$

Where $z_t$ is an intermediate vector, $W_{Res3}$ is the projection matrix.

Since fast LSTM is faster to train, we also examine these thoughts into fast LSTM and test their performance on TIMIT, THCHS-30, Librispeech and Switchboard tasks.

## IV. EXPERIMENTS

We use nnet3 recipes in Kaldi, a popular toolkit for speech recognition, to implement our LSTM architectures. The input of neural networks is a 300-dimensional vector, consisting of 40 dimensional MFCC (Mel Frequency Cepstral Coefficient) features without cepstral truncation of five frames, which are spliced across in ±2 frames of context, and 100 dimensional i-vectors to perform speaker adaptation [20].

On TIMIT and THCHS-30 tasks, all the implemented LSTM architectures have 2, 3 or 4 LSTM layers followed by a fully connected layer and a softmax layer, while the depth is 3 or 6 on Librispeech and Switchboard tasks. They are trained by cross-entropy (CE) criterion [21]. We use several Nvidia GPUs to accelerate the training based on the parallel training algorithm proposed in [22].

In order to evaluate our recipes quickly, speed perturbation [23] is not considered in our system. Speed perturbation is adopted to generate more training data for audio augmentation in Kaldi by default, but its performance improvement is very limited.

Since the training of LSTM is very slow, LSTM is just operated on TIMIT task. And these thoughts with fast LSTM are evaluated on THCHS-30, Librispeech and Switchboard tasks.

### A. On TIMIT Task

TIMIT speech corpus is widely used in acoustic-phonetic studies. In our experiments, 3.14 hours of data was selected as training data, with 0.16 hours as core test set and 0.81 hours as a complete test set, which was suggested by official [24]. The core test set is the subset of the complete test set.

The baseline LSTM implemented on TIMIT is the projected LSTM provided by Kaldi and Residual LSTM [6]. All LSTM, fast LSTM and their residual architectures have 1024-dimensional cell, with 512-dimensional recurrent projection and zero-dimensional no-recurrent projection, i.e., all elements of $y_t$ is used for recurrence. "Residual LSTM" is the architecture proposed in [6]. All neural networks are trained for 15 epochs for the amount of training data is very small. The results are illustrated in Table I.

TABLE I. RESULTS OF TIMIT TASK (PER%)

| AM Types | #D [a] | #P [b] | Core Test | Complete Test |
|---|---|---|---|---|
| LSTM | 2 | 9.6M | 21.0 | 20.3 |
|  | 3 | 14.3M | 21.2 | **20.0** |
|  | 4 | 19.0M | **20.8** | 20.1 |
| Residual LSTM | 2 | 9.8M | 21.4 | 20.4 |
|  | 3 | 14.6M | 20.8 | 20.1 |
|  | 4 | 19.3M | **20.7** | **19.7** |
| LSTM Res-1 | 2 | 12.5M | 20.0 | 19.1 |
|  | 3 | 18.8M | 20.0 | 18.7 |
|  | 4 | 25.1M | **19.3** | **18.4** |
| LSTM Res-2 | 2 | 10.0M | 21.2 | 19.8 |
|  | 3 | 15.0M | **19.8** | 19.3 |
|  | 4 | 20.0M | 20.2 | **19.1** |
| LSTM Res-3 | 2 | 10.5M | 20.1 | 19.1 |
|  | 3 | 15.8M | **19.9** | 18.8 |
|  | 4 | 21.0M | 20.2 | **18.6** |
| Fast LSTM | 2 | 9.6M | 20.3 | 19.3 |
|  | 3 | 14.3M | 20.1 | 18.7 |
|  | 4 | 19.0M | **20.0** | **18.7** |
| Fast LSTM Res-1 | 2 | 12.5M | 19.5 | 18.5 |
|  | 3 | 18.8M | **19.2** | **18.1** |
|  | 4 | 25.1M | 19.5 | **18.1** |
| Fast LSTM Res-2 | 2 | 10.0M | 19.9 | 18.6 |
|  | 3 | 15.0M | **19.3** | 18.2 |
|  | 4 | 20.0M | 19.6 | **17.9** |
| Fast LSTM Res-3 | 2 | 10.5M | 19.6 | 18.7 |
|  | 3 | 15.8M | **19.2** | 18.2 |
|  | 4 | 21.0M | 19.3 | **18.0** |

a. #D means depth of the neural network, i.e., the number of LSTM layers.
b. #P is the number of parameters.

According to the matrixes $W_{ic}$, $W_{fc}$ and $W_{oc}$ in (1), (2) and (5), which are all diagonal matrices and have very small number of parameters when compared them with other general matrices such as $W_{ix}$ and $W_{ir}$, the number of parameters of LSTM is approximately equal to fast LSTM.

As shown in Table I, Residual LSTM proposed in [6] has a very small improvement in performance; the residual methods we proposed outperform both baseline LSTM and

Residual LSTM, with a relative reduction in PER about 8% on complete test set for LSTM Res-1. Besides, fast LSTM is not only faster than standard LSTM, but also outperforms standard LSTM.

When applied to fast LSTM, our residual recipes also gain performance improvement, with a relative reduction in PER about 4% on both core test set and complete test set. More importantly, our fast LSTM Res recipes have the best performance. So, on the following THCHS-30 task, we just apply these three ideas to fast LSTM.

*B. On THCHS-30 Task*

THCHS-30 (Tsinghua Chinese 30-hour database) is a free and open-source Chinese speech database. On this task, we use about 25 hours of data for training and about 6 hours of data for testing [14].

Different with TIMIT task, the baseline fast LSTM and the three residual fast LSTM recipes here are implemented with 1024-dimensional cell, 256-dimensional recurrent projection and 256-dimensional no-recurrent projection. Besides, we train all neural networks for 8 epochs. When decoding, we perform both phone decoding and word decoding, which convert the speech signal into phone sequences and word sequences. The results are shown in Table II.

TABLE II. RESULTS OF THCHS-30 TASK (WER%/PER%)

| AM Types | #D | #P | Word Task [a] | Phone Task |
|---|---|---|---|---|
| Fast LSTM | 2 | 8.2M | 22.94 | 9.53 |
| | 3 | 11.9M | 22.74 | 9.15 |
| | 4 | 15.6M | **22.62** | **9.01** |
| Fast LSTM Res-1 | 2 | 11.2M | 23.29 | 9.13 |
| | 3 | 16.4M | 22.97 | 8.87 |
| | 4 | 21.7M | **22.83** | **8.74** |
| Fast LSTM Res-2 | 2 | 8.6M | 22.64 | 9.31 |
| | 3 | 12.6M | 22.36 | 8.75 |
| | 4 | 16.5M | **22.15** | **8.38** |
| Fast LSTM Res-3 | 2 | 9.2M | 22.92 | 8.88 |
| | 3 | 13.4M | 22.57 | 8.72 |
| | 4 | 17.6M | **22.40** | **8.31** |

a. WER stands for Word Error Rate on word task, and PER is for phone task.

Obviously, the recipes we proposed outperform the standard fast LSTM in most cases, especially on phone task. And in several cases, there is about 7% relative improvement of performance on phone task and about 2% relative improvement of performance on word task.

Meanwhile, the best result provided by CSLT@THU is 23.18% on word task and 10.01% on phone task [25]. And our best result is 22.15% on word task and 8.31% on phone task, with 4.4% relative reduction in WER and 16.9% relative reduction in PER.

*C. On Librispeech Task*

Librispeech corpus is a free and open-source read speech data set [15]. In this paper, we use train-clean-100 as training data, which contains about 100-hour training data. And the performance is evaluated on dev-clean, test-clean, dev-other and test-other sets, among which "other" series are more complex than "clean" series, leading a worse performance on "other" series.

The networks trained here have three LSTM layers and a softmax layer, with 256-dimensional recurrent projection and 256-dimensional no-recurrent projection. Only fast LSTM Res-1 and fast LSTM are trained for 15 epochs. When decoding, we use a small 3-gram language model to decode firstly and then use a bigger 4-gram language model to rescore. The rescored results on four sets are shown in Table III.

TABLE III. RESULTS OF LIBRISPEECH TASK (WER%)

| AM Types | #D | dev clean | test clean | dev other | test other |
|---|---|---|---|---|---|
| Fast LSTM | 3 | 6.70 | 7.47 | 25.09 | 26.67 |
| Fast LSTM Res-1 | 3 | **5.74** | **6.35** | **21.39** | **22.58** |

As shown in Table III, the fast LSTM Res-1 method proposed by us outperforms the standard fast LSTM, with more than 14% relative reduction in WER on all dev and test sets.

*D. On Switchboard Task*

Switchboard is a collection of about 2,400 two-sided telephone conversations among 543 speakers from all areas of the United States [16]. We use Switchboard-1 Release 2 for training, and when training we remove excess utterances once they appear more than 300 times with the same transcription.

2000 HUB5 English Evaluation Set (Eval2000) [26] is used for testing, which consists of 20 unreleased telephone conversations from the Switchboard studies and 20 telephone conversations from Callhome American English Speech. When decoding, we compute WER on the whole evaluation set, Switchboard subset and Callhome subset.

Similar to Librispeech task, the networks implemented here all have 1024-dimensional cell, with 256-dimensional recurrent projection and 256-dimensional no-recurrent projection. We train the 6-layer neural networks for 8 epochs. Only the Res-1 method is implemented here. When decoding, we firstly use a 3-gram language model computing from Switchboard corpus to decode, then we use a 4-gram language model computing from Switchboard corpus and Fisher corpus [27] to rescore. The rescored results are shown in Table IV.

TABLE IV. RESULTS OF SWITCHBOARD TASK (WER%)

| AM Types | #D | Eval2000 | Switchboard subset | Callhome subset |
|---|---|---|---|---|
| Fast LSTM | 6 | 17.2 | 11.4 | 22.8 |
| Fast SLTM Res-1 | 6 | **16.8** | **11.4** | **21.9** |

As shown in the above table, the fast LSTM Res-1 method outperforms the standard fast LSTM on Callhome subset, with about 4% reduction in WER, which leads to about 2% reduction in WER on the whole Eval2000 set.

Meanwhile, the performance on Switchboard subset remains the same, which means the improved residual LSTM proposed by us achieves stronger generalization performance.

## V. CONCLUSIONS

In this paper, we propose an improved residual LSTM architecture for acoustic modeling. The experiments on TIMIT corpus show that our recipes outperform the standard LSTM and the Residual LSTM proposed in [6], with about 8% reduction in PER on complete test set. Also, we apply our thoughts to fast LSTM and achieve our best performance on TIMIT, THCHS-30, Librispeech and Switchboard corpora. Besides, the best performance ever was achieved by using our fast LSTM structure on THCHS-30 task. The experiments on Switchboard also show that the improved residual LSTM has stronger generalization ability.

Next, we are interested in bidirectional LSTM [28] and deeper LSTM. We would also like improving performance on other corpora by using our improved residual LSTM architectures.

ACKNOWLEDGMENT

This work is partially supported by the National Natural Science Foundation of China (Nos. 11590770-4).

REFERENCES

[1] F. Seide, G. Li, and D. Yu, "Conversational Speech Transcription Using Context-Dependent Deep Neural Networks," Proc. Interspeech, Aug. 2011, pp. 437-440, doi: 10.1.1.368.3047.

[2] Y. Bengio, P. Simard and P. Frasconi, "Learning long-term dependencies with gradient descent is difficult," IEEE Transactions on Neural Networks, vol. 5, no. 2, pp. 157-166, Mar. 1994, doi: 10.1109/72.279181.

[3] S. Hochreiter and J. Schmidhuber, "Long Short-Term Memory," Neural Computation, vol. 9, no. 8, pp. 1735-1780, Nov. 15 1997, doi: 10.1162/neco.1997.9.8.1735.

[4] H. Sak, S. Andrew, and B. Françoise. "Long short-term memory recurrent neural network architectures for large scale acoustic modeling," Proc. Interspeech, Sep. 2014, pp. 338-342.

[5] K. He, X. Zhang, S. Ren and J. Sun, "Deep Residual Learning for Image Recognition," 2016 IEEE Conference on Computer Vision and Pattern Recognition (CVPR), Las Vegas, NV, 2016, pp. 770-778, doi: 10.1109/CVPR.2016.90.

[6] J. Kim, M. El-Khamy, and J. Lee, "Residual LSTM: Design of a Deep Recurrent Architecture for Distant Speech Recognition," arXiv preprint arXiv:1701.03360, Jan. 2017.

[7] G. Saon, G. Kurata, T. Sercu, K. Audhkhasi, S. Thomas, D. Dimitriadis, X. Cui, B. Ramabhadran, M. Picheny, LL. Lim, and B. Roomi, "English conversational telephone speech recognition by humans and machines," arXiv preprint arXiv:1703.02136, Mar. 2017.

[8] W. Xiong, J. Droppo, X. Huang, F. Seide, M. Seltzer, A. Stolcke, D. Yu, and G. Zweig, "Achieving human parity in conversational speech recognition," arXiv preprint arXiv:1610.05256, Oct. 2016.

[9] W. Xiong, J. Droppo, X. Huang, F. Seide, M. Seltzer, A. Stolcke, D. Yu, and G. Zweig, "The Microsoft 2016 conversational speech recognition system," arXiv preprint arXiv:1609.03528, Sep. 2016.

[10] Y. Zhang, W. Chan, and N. Jaitly, "Very deep convolutional networks for end-to-end speech recognition," arXiv preprint arXiv:1610.03022, Oct. 2016.

[11] S.P.P. Selvaraj, and S. Konam, "Deep Learning for Speaker Recognition," unpublished, available online, accessed Apr. 19, 2017. https://skonam.github.io/course_projects/10701.pdf.

[12] O. Ogunmolu, X. Gu, S. Jiang, and N. Gans, "Nonlinear Systems Identification Using Deep Dynamic Neural Networks," arXiv preprint arXiv:1610.01439, Oct. 2016.

[13] J.S. Garofolo, L.F. Lamel, W.M. Fisher, J.G. Fiscus, D.S. Pallett, N.L. Dahlgren, and V. Zue, "Timit acoustic-phonetic continuous speech corpus," Linguistic Data Consortium, Philadelphia, 1993.

[14] D. Wang, X. Zhang, "THCHS-30: A free Chinese speech corpus," arXiv preprint arXiv:1512.01882, Dec. 2015.

[15] V. Panayotov, G. Chen, D. Povey and S. Khudanpur, "Librispeech: An ASR corpus based on public domain audio books," 2015 IEEE International Conference on Acoustics, Speech and Signal Processing (ICASSP), South Brisbane, QLD, 2015, pp. 5206-5210, doi: 10.1109/ICASSP.2015.7178964.

[16] J. Godfrey, and E. Holliman, "Switchboard-1 Release 2 LDC97S62," Web Download, Philadelphia: Linguistic Data Consortium, 1993.

[17] K. Kawakami, "Supervised Sequence Labelling with Recurrent Neural Networks", Ph.D. thesis, Technical University of Munich, 2008.

[18] D. Povey, A. Ghoshal, G. Boulianne, L. Burget, O. Glembek, N. Goel, M. Hannemann, P. Motlicek, Y. Qian, P. Schwarz, and J. Silovsky, "The Kaldi speech recognition toolkit," 2011 IEEE workshop on automatic speech recognition and understanding (ASRU), Dec. 2011.

[19] Y. Miao, J. Li, Y. Wang, S. X. Zhang and Y. Gong, "Simplifying long short-term memory acoustic models for fast training and decoding," 2016 IEEE International Conference on Acoustics, Speech and Signal Processing (ICASSP), Shanghai, 2016, pp. 2284-2288, doi: 10.1109/ICASSP.2016.7472084.

[20] G. Saon, H. Soltau, D. Nahamoo and M. Picheny, "Speaker adaptation of neural network acoustic models using i-vectors," 2013 IEEE Workshop on Automatic Speech Recognition and Understanding (ASRU), Olomouc, Dec. 2013, pp. 55-59. doi: 10.1109/ASRU.2013.6707705.

[21] P.T. De Boer, D.P. Kroese, S. Mannor, and R.Y. Rubinstein, "A tutorial on the cross-entropy method," Annals of operations research, Feb. 2005, Volume 134, Issue 1, pp 19–67, doi: 10.1007/s10479-005-5724-z.

[22] D. Povey, X. Zhang, and S. Khudanpur, "Parallel training of Deep Neural Networks with Natural Gradient and Parameter Averaging," 2015 International Conference on Learning Representations (ICLR), May. 2015.

[23] T. Ko, V. Peddinti, D. Povey, and S. Khudanpur, "Audio augmentation for speech recognition," Proc. Interspeech, Germany, Sep. 2015.

[24] C. Lopes, and F. Perdigao, "Phone recognition on the TIMIT database," Speech Technologies/Book, vol. 1, pp. 285-302, 2011.

[25] THCHS-30 CHALLENGE, Accessed April 19, 2017. http://data.cslt.org/thchs30/challenges/asr.html.

[26] Linguistic Data Consortium, "2000 HUB5 English Evaluation Speech LDC2002S09," Web Download, Philadelphia: Linguistic Data Consortium, 2002.

[27] C. Cieri, D. Miller, and K. Walker, "The Fisher Corpus: A Resource for the Next Generations of Speech-to-Text," Tenth International Conference on Language Resources and Evaluation (LREC), vol. 4, pp. 69-71, 2004.

[28] A. Graves, N. Jaitly and A. Mohamed, "Hybrid speech recognition with Deep Bidirectional LSTM," 2013 IEEE Workshop on Automatic Speech Recognition and Understanding (ASRU), Olomouc, 2013, pp. 273-278, doi: 10.1109/ASRU.2013.6707742.